\documentclass[lettersize,journal]{IEEEtran}
\usepackage{amsmath,amsfonts}
\usepackage{algorithmic}
\usepackage{algorithm}
\usepackage{array}
\usepackage{hyperref}
\usepackage[caption=false,font=normalsize,labelfont=sf,textfont=sf]{subfig}
\usepackage{booktabs}
\usepackage{multirow}
\usepackage{caption}
\usepackage{textcomp}
\usepackage{stfloats}
\usepackage{url}
\usepackage{verbatim}
\usepackage{graphicx}
\usepackage{subcaption} 
\usepackage{cite}
\usepackage{xcolor}
\usepackage{pifont}
\usepackage{orcidlink}
\usepackage{enumitem}
\newcommand{\eg}{\textit{e}.\textit{g}.}
\newcommand{\ie}{\textit{i}.\textit{e}.}

\newcommand{\cmark}{\ding{51}}
\newcommand{\xmark}{\ding{55}}

\newcounter{bxincomm}
\definecolor{aqua}{rgb}{0.00,0.67,0.80}

\newcounter{todocomm}

\begin{document}

\title{Retrieval-Augmented Multimodal Learning for Enzyme–Substrate Interaction Prediction Under Low-Homology Shift
}

\author{Chen~Liu,
        Bingxin~Zhou\orcidlink{0000-0002-3897-9766},
        Xinyuan~Wang,
    Ming~Li\orcidlink{0000-0002-1218-2804},
        Guisheng~Fan\orcidlink{0000-0002-2702-0242},
        Liang~Hong\orcidlink{0000-0003-0107-336X}

\thanks{This work was supported in part by the National Natural Science
Foundation of China under Grant No.62302291. Chen Liu and Bingxin Zhou contributed equally to this work. Corresponding author: 
Bingxin Zhou \textit{(bingxin.zhou@sjtu.edu.cn)}.}
\IEEEcompsocitemizethanks{
\IEEEcompsocthanksitem Chen~Liu and Guisheng~Fan are with School of Information Science and Engineering, East China University of Science and Technology, Shanghai 200237, China.
\IEEEcompsocthanksitem Bingxin Zhou and Liang Hong are with Institute of Natural Sciences and Zhangjiang Institute for Advanced Study, Shanghai Jiao Tong University, Shanghai 200240, China. 
\IEEEcompsocthanksitem Xinyuan Wang is with School of Life Sciences and Biotechnology, Shanghai Jiao Tong University. Shanghai 200240, China.
\IEEEcompsocthanksitem Ming Li is with Zhejiang Key Laboratory of Intelligent Education Tech-
nology and Application, Zhejiang Normal University, Jinhua, China.
}
}

\markboth{Submitted to IEEE Transactions On Knowledge And Data Engineering
}%
{Liu and Zhou \MakeLowercase{\textit{et al.}}: RAMMESI for Enzyme-Substract Interaction Prediction}


\maketitle

\begin{abstract}
Enzyme–substrate interaction (ESI) prediction is a fundamental computational task for biocatalyst discovery and reaction screening in large biochemical spaces. In practical settings, ESI prediction is challenged by sparse positive supervision and low-homology distribution shift, where test enzymes share limited sequence identity with those observed during training. To address these challenges, we propose RAMMESI, a retrieval-augmented multimodal framework for robust ESI prediction. RAMMESI learns explicit pairwise enzyme–substrate representations through directional cross-modal interaction modeling and adaptive fusion. To enhance robustness, RAMMESI retrieves neighboring enzymes at inference time, recombines them with the query substrate, and aggregates the resulting pairwise predictions as contextual evidence. To improve learning under sparse positive supervision, we further adopt an imbalance-aware weighted-BCE objective. Experiments on two ESI benchmarks under sequence-identity-aware splits demonstrate that RAMMESI achieves consistently strong performance, with particular advantages in more challenging low-identity regimes. In addition, the retrieval module improves multiple ESI backbones in a plug-and-play manner, suggesting that retrieval provides a general mechanism for improving robustness under homology shift. The source code is publicly available at the following link \url{https://github.com/code4luck/RAMMESI.git}
\end{abstract}


\section{Introduction}
\IEEEPARstart{E}{nzyme}-substrate interaction (ESI) prediction is a heterogeneous pairwise prediction task that evaluates whether a given enzyme and substrate are likely to form a reactive pair. It is a fundamental computational task for biocatalyst discovery \cite{chapman2018industrial} and industrial bioprocessing \cite{buller2023nature}, as it enables large-scale prioritization of reactive enzyme-substrate candidates before costly experimental validation. Unlike enzyme-oriented functional annotation, which predicts enzyme function without specifying a candidate substrate \cite{yu2023enzyme}, ESI prediction evaluates whether a given enzyme-substrate pair is likely to be reactive. Since experimentally verified enzyme-substrate pairs remain limited relative to the large candidate space, accurate ESI prediction is essential for scalable reaction screening \cite{bozkurt2025accelerating,uniprot2025uniprot}.

\begin{figure}[!t]
\centering
{\includegraphics[width=\linewidth]{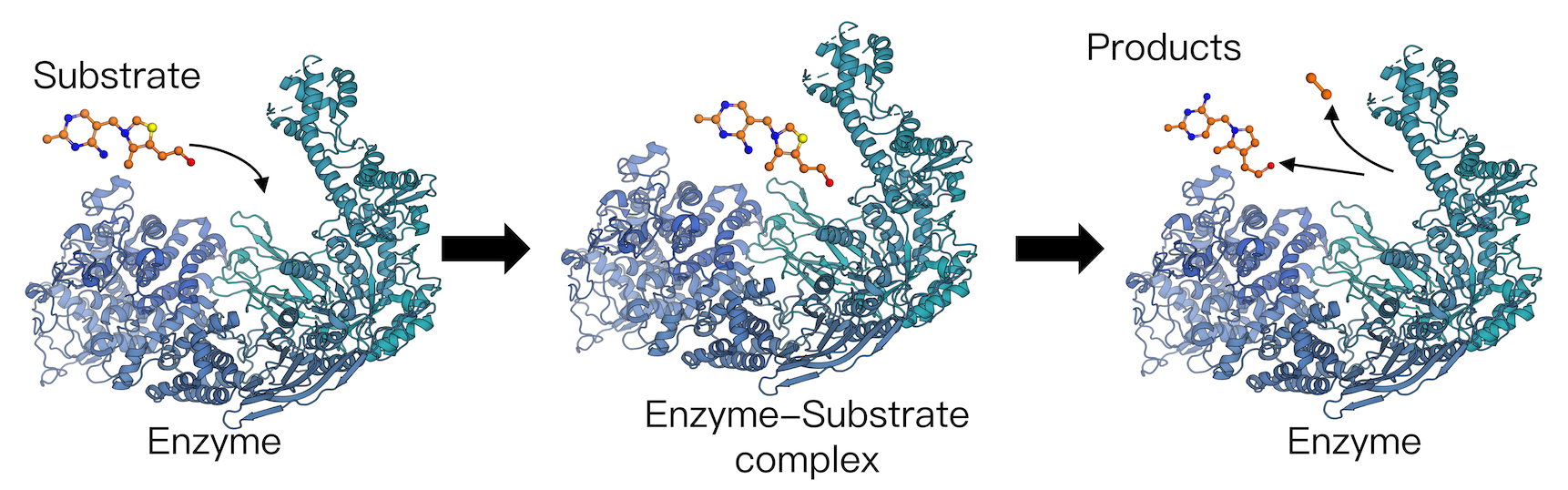}}
\caption{Schematic illustration of substrate binding, complex formation, and product release (left to right).}
\label{fig_esi_comp}
\end{figure}

Early studies on ESI prediction mainly relied on conventional machine learning models and expert-designed features, such as physicochemical properties, sequence-derived descriptors, and homology-based signals \cite{rottig2010combining,pertusi2017predicting,yang2018functional,mou2021machine}. These approaches enabled more scalable screening than wet-lab experiments, but their reliance on expert knowledge limited their ability to capture complex compatibility patterns and generalize beyond specific enzyme families or functional categories \cite{banerjee2022enzymclass,kroll2023general}. More recently, advances in deep learning have improved biomolecular interaction modeling across molecular prediction tasks \cite{li2023giant,ma2022kg,ma2024learning,chen2025local,bi2025ssppi}, while pretrained biomolecular models have enabled general-purpose representation learning from protein and molecular inputs \cite{ji2024exploring,chen2025protein,chen2025accurate}. Recent ESI predictors therefore increasingly employ pretrained encoders with learnable fusion or prediction modules, reducing dependence on hand-crafted features and achieving stronger predictive performance \cite{kroll2023general,du2025fusionesp}.

Despite these advances, current methods still insufficiently model the pairwise compatibility underlying substrate-specific enzyme reactivity. \textbf{A key modeling challenge is to effectively fuse enzyme and substrate representations for substrate-conditioned reactivity prediction}. This challenge arises because enzyme-centric functional characterization alone is insufficient to determine the reactivity of a specific enzyme-substrate pair \cite{zhang2025sefp}. Enzyme reactivity can vary substantially across substrates, and some enzymes exhibit broad substrate specificity \cite{hult2007enzyme,wang2024multi}. Although pairwise ESI predictors address substrate-conditioned prediction more directly, they often rely on simple concatenation or shallow fusion of protein and substrate embeddings, which may fail to capture directional and context-dependent compatibility patterns \cite{campbell2024viper,du2025fusionesp}. As Fig.~\ref{fig_esi_comp} illustrates, substrate binding and complex formation involve coordinated enzyme-substrate recognition, further motivating cross-modal interaction modeling beyond independent encoding and late-stage fusion.

\begin{figure*}[!t]
\centering
{\includegraphics[width=\linewidth]{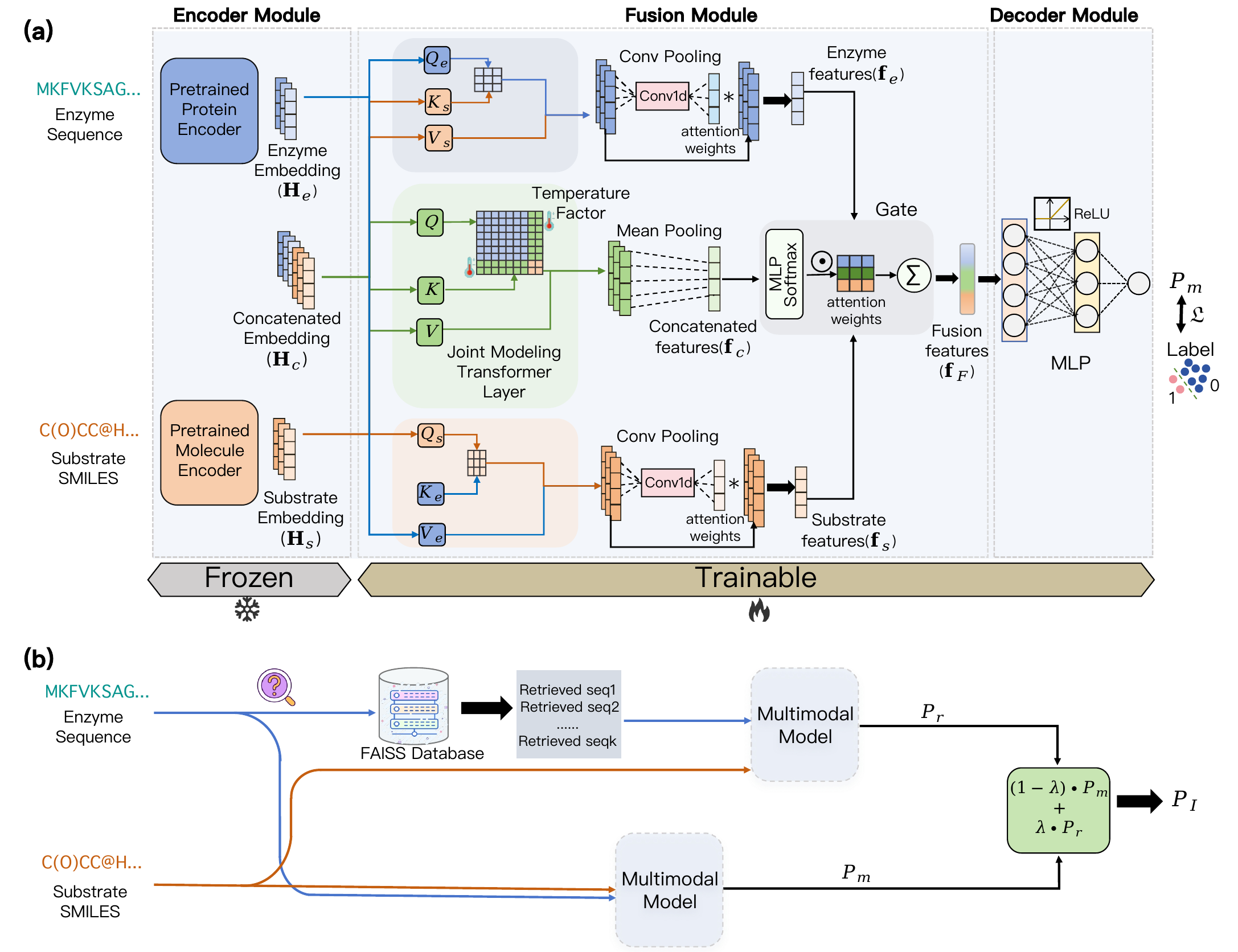}}
\caption{Overview of the proposed RAMMESI framework. (a) RAMMESI encodes enzyme and substrate inputs, models their cross-modal interactions, adaptively fuses complementary representations, and outputs the interaction probability. (b) During inference, RAMMESI retrieves neighboring enzymes in the embedding space, pairs them with the query substrate to get retrieval predictions, and aggregates them with the query prediction to produce the final interaction probability ($P_I$).}
\label{fig_frame}
\end{figure*}

Beyond modeling limitations, ESI prediction also faces two training and evaluation challenges. First, \textbf{reliable generalization to low-homology enzymes remains difficult}. Predictions for close homologs can often exploit sequence similarity, whereas low-homology enzymes provide weaker evidence for annotation transfer and better reflect the need to predict reactivity beyond well-characterized enzyme families. In practice, this challenge is evaluated through homology-aware sequence-identity splits, where test enzymes share limited identity with training enzymes and may differ in structure, active-site context, or substrate preference \cite{bileschi2022using,paton2025connecting,tan2026venusx}. Existing methods may degrade substantially under such distribution shifts \cite{kroll2023general}. Second, \textbf{learning to recover rare reactive pairs under severe class imbalance remains challenging}. Verified reactive pairs are much fewer than non-reactive candidates, causing standard cross-entropy training to be dominated by negative examples. This imbalance can bias prediction toward the majority class and reduce recall for true interacting pairs~\cite{hua2024reactzyme,uniprot2025uniprot}.

Effective ESI prediction requires stronger pairwise interaction modeling, robustness under homology-induced distribution shift, and reliable learning from imbalanced supervision. To this end, we propose RAMMESI, a retrieval-augmented multimodal framework for ESI prediction. RAMMESI treats ESI as a fine-grained pairwise prediction problem over protein and small-molecule inputs, and addresses the above challenges through directional cross-modal interaction modeling, adaptive fusion of complementary interaction views, imbalance-aware optimization, and inference-time retrieval augmentation (Fig.~\ref{fig_frame}). Our main contributions are summarized as follows.
\begin{enumerate}[leftmargin=*]
\item \textbf{Substrate-conditioned pairwise multimodal modeling.} 
We propose RAMMESI to model ESI as substrate-conditioned pairwise reactivity prediction. It captures bidirectional enzyme-to-substrate and substrate-to-enzyme interaction views and adaptively fuses them with joint pair-level context for fine-grained prediction.
\item \textbf{Support-pair retrieval for low-homology generalization.}
We introduce an inference-time retrieval augmentation that transforms contextual evidence from neighboring enzymes into substrate-specific support pairs and improves reliability for low-sequence-identity test enzymes.
\item \textbf{Imbalance-aware optimization for rare reactive pairs.}
We formulate ESI training under a weighted-BCE framework and introduce threshold-aware smooth weighting to reduce the dominance of easy negatives under sparse positive supervision.
\end{enumerate}

The rest of this paper is organized as follows. Section~\ref{real_work} reviews related studies on enzyme–substrate interaction prediction and retrieval augmentation. Section~\ref{method} presents the formulation and the details of the RAMMESI framework. Section~\ref{exp} reports the experimental settings and various experiments to validate
the effectiveness of RAMMESI. Finally, Section~\ref{conclusion} concludes this paper.

\section{Related work}
\label{real_work}
In this section, we review closely related literature
to our work, including enzyme–substrate interaction prediction and retrieval augmentation.

\emph{Enzyme-Substrate Interaction Prediction}:
Existing computational studies on ESI can be broadly categorized into enzyme-centric functional prediction and pairwise enzyme-substrate prediction. Enzyme-centric methods (\eg, EC-number assignment) provide useful coarse functional labels but do not directly determine whether a specific enzyme-substrate pair is likely to be reactive \cite{dhanuka2023comprehensive,kulmanov2024protein,tan2025venusfactory}. More recent ESI studies have increasingly moved toward pairwise prediction by leveraging protein and molecular representations learned from pretrained encoders \cite{qian2024deep,du2025fusionesp,nie2025omniesi}. Within this direction, existing methods can be further grouped into two-stage pipelines and end-to-end models. Two-stage methods first obtain enzyme and molecular representations and then train a separate classifier. For example, ESP \cite{kroll2023general} fine-tunes pretrained representations before training a separate classifier for ESI prediction. End-to-end approaches jointly learn representations and predictions within a unified architecture, using attention-based feature fusion \cite{campbell2024viper,nie2025omniesi} or representation alignment \cite{du2025fusionesp} to model cross-modal enzyme-substrate interactions more explicitly. In addition to sequence-based modeling, enzyme structural information can further benefit ESI prediction when reliable structures are available, as exemplified by EZSpecificity~\cite{cui2025enzyme}; however, obtaining experimentally resolved structures remains costly and labor-intensive, which can limit the broader use of structure-dependent methods. Overall, existing ESI methods have increasingly moved toward pairwise substrate-conditioned prediction with learnable multimodal representations. RAMMESI follows this end-to-end pairwise prediction paradigm and focuses on explicit ESI modeling for fine-grained reactivity prediction.

\emph{Retrieval Augmentation}:
Retrieval augmentation couples a parametric model with external exemplar evidence and has emerged as a general strategy for improving prediction beyond purely parametric inference \cite{lewis2020retrieval,fan2024survey}. Although originally developed in retrieval-augmented generation, many retrieval-based frameworks incorporate retrieved examples during training, for example, by augmenting model inputs, constructing retrieval-conditioned objectives, or improving representation learning with neighbor-derived context \cite{weitzman2025protriever,li2025large}. Inference-time retrieval has also been explored in general prediction settings through exemplar-supported aggregation, where predictions are adjusted using retrieved neighbors without retraining the model \cite{gao2023retrieval,peng2025graph}. In biomolecular modeling, however, retrieval-based strategies are still primarily used to provide training-time context or single-modality protein evidence, often through homology search or embedding-space lookup \cite{zhang2025deep,notin2022tranception,datta2025embedding}. For example, VenusREM \cite{tan2025venusrem} constructs position-specific evolutionary logits and linearly interpolates them with protein language model outputs to inject evolutionary priors at test time. Overall, inference-time retrieval remains less explored in biomolecular prediction, especially for pairwise multimodal ESI frameworks. In particular, inference-time single-side neighborhood retrieval with neighbor-supported aggregation remains underexplored for improving robustness under low-homology ESI evaluation.

\section{Methodology}
\label{method}
This section presents RAMMESI. We first provide a framework overview, followed by its main components for representation learning, interaction modeling, prediction, imbalance-aware optimization, and retrieval-augmented inference.

\subsection{Overview}
We formulate ESI prediction as a pairwise binary classification task. 
Let $e$ denote an enzyme sequence, $s$ denote a substrate SMILES string, and 
$y\in\{0,1\}$ denote the binary interaction label, where $y=1$ indicates an interacting pair. 
Given a pair $(e,s)$, RAMMESI learns a predictor 
$\psi_\Theta(\cdot)$ and outputs the probability 
$P_m=\psi_\Theta(e,s) \in (0,1)$ that the pair is reactive. During training (Fig.~\ref{fig_frame}(a)), the multimodal model is optimized end-to-end with an imbalance-aware objective $\mathcal{L}$ to learn pairwise interaction patterns. During inference (Fig.~\ref{fig_frame}(b)), the trained model uses a retrieval module to incorporate neighborhood evidence from similar enzymes and improve robustness under low-homology evaluation.

\subsection{The RAMMESI Model Architecture}
As shown in Fig.~\ref{fig_frame}(a), RAMMESI consists of three modules: an encoder, a fusion module, and a decoder. The encoder extracts representations from enzyme sequences and substrate molecules. The fusion module models cross-modal interactions and integrates complementary interaction views. The decoder maps the fused representation to the final interaction probability. We describe these modules below.

\subsubsection{Encoder Module}
The encoder module obtains representations for enzyme sequences and substrate molecules. Enzyme sequences are encoded by a pretrained protein language model into $\mathbf{H}_e \in \mathbb{R}^{L_e \times d}$, and substrate SMILES strings are encoded by a pretrained molecular encoder into $\mathbf{H}_s \in \mathbb{R}^{L_s \times d}$, where $L_e$ and $L_s$ denote the embedding lengths of the enzyme and substrate, respectively, and $d$ is the encoder embedding dimension. The resulting enzyme and substrate embeddings are then passed to the fusion module for interaction modeling. During training, both encoders are kept frozen to reduce optimization cost and preserve stable pretrained representations.

\subsubsection{Fusion Module}The fusion module (Fig.~\ref{fig_frame}(a) Fusion Module) aims to explicitly model cross-modal enzyme–substrate interactions and transform variable-length enzyme and substrate embeddings into discriminative pairwise features for downstream prediction. To this end, we design the fusion module around three complementary views: directional cross-modal interaction modeling, joint complex-level modeling, and adaptive multi-view fusion.

\paragraph{Directional Cross-modal Interaction Modeling}
Enzymes and substrates play asymmetric roles in the reaction process. To preserve this asymmetry, we use cross-attention to construct directional interaction views. Given $\mathbf{Q} \in \mathbb{R}^{L_Q\times d_a}$ and $\mathbf{K}, \mathbf{V} \in \mathbb{R}^{L_K \times d_a}$, where  $L_Q$ denotes the length of the query representation, $L_K$ denotes the length of the key and value representations, and $d_a$ is the attention hidden dimension, cross-attention is computed as
\begin{align}
\mathbf{R} &= \frac{\mathbf{Q}\mathbf{K}^{\top}}{\sqrt{d_a}} \in \mathbb{R}^{L_Q \times L_K}, \\
\mathbf{A} &= \rm softmax(\mathbf{R}) \in \mathbb{R}^{L_Q \times L_K}, \\
\mathbf{Z} &= \mathbf{A}\mathbf{V} \in \mathbb{R}^{L_Q \times d_a},
\end{align}
where $\mathbf{R}$ denotes the attention logits and $\mathbf{A}$ denotes the normalized attention weights. Since each output vector $\mathbf{z}_i$ in $\mathbf{Z} \in \mathbb{R}^{L_Q \times d_a}$ corresponds to a query token, cross-attention is directional with respect to the query choice. Accordingly, RAMMESI uses two complementary branches: an enzyme-to-substrate branch with $\mathbf{H}_e$ as queries and $\mathbf{H}_s$ as keys and values, and a substrate-to-enzyme branch with the roles reversed. The outputs of both branches are retained as complementary cross-modal interaction views.

\paragraph{Joint Complex-level Modeling}
To capture global context at the enzyme-substrate complex level, we further introduce a joint transformer branch. Using the encoder outputs $\mathbf{H}_e$ and $\mathbf{H}_s$, we concatenate them as
\begin{equation}
\mathbf{H}_c = [\mathbf{H}_e; \mathbf{H}_s] \in \mathbb{R}^{L_c \times d}, \qquad L_c = L_e + L_s,
\end{equation}
where $L_c$ denotes the length of concatenated enzyme and substrate embeddings, and compute self-attention over the joint sequence:
\begin{equation}
\mathbf{Q}_c = \mathbf{H}_c \mathbf{W}_Q,\quad
\mathbf{K}_c = \mathbf{H}_c \mathbf{W}_K,\quad
\mathbf{V}_c = \mathbf{H}_c \mathbf{W}_V,
\end{equation}
\begin{equation}
\mathbf{R}_c = \frac{\mathbf{Q}_c \mathbf{K}_c^\top}{\sqrt{d_a}} \in \mathbb{R}^{L_c \times L_c}.
\end{equation}

Enzyme and substrate embeddings often differ in length, and joint attention may dilute cross-modal signals over the longer modality. To mitigate this, we decompose the attention logits into intra-modal and cross-modal blocks:
\begin{equation}
\mathbf{R}_c =
\begin{bmatrix}
\mathbf{R}_{ee} & \mathbf{R}_{es} \\
\mathbf{R}_{se} & \mathbf{R}_{ss}
\end{bmatrix},
\end{equation}
where $\mathbf{R}_{ee}$ and $\mathbf{R}_{ss}$ denote intra-modal logits, and $\mathbf{R}_{es}$ and $\mathbf{R}_{se}$ denote cross-modal logits. We then apply a temperature factor $\tau$ to the cross-modal blocks:
\begin{equation}
\tilde{\mathbf{R}}_c =
\begin{bmatrix}
\mathbf{R}_{ee} & \tau \mathbf{R}_{es} \\
\tau \mathbf{R}_{se} & \mathbf{R}_{ss}
\end{bmatrix},
\end{equation}
followed by standard attention normalization:
\begin{align}
\mathbf{A}_c &= \rm softmax(\tilde{\mathbf{R}}_c), \\
\mathbf{Z}_c &= \mathbf{A}_c \mathbf{V}_c.
\end{align}
Here, $\tau$ calibrates the contribution of cross-modal alignment before softmax normalization.

\paragraph{Adaptive Multi-view Fusion}
The transformed representations of the three interaction branches are mapped into three view-specific feature vectors $\{\mathbf{f}_e,\mathbf{f}_c,\mathbf{f}_s\} \in \mathbb{R}^{d_a}$ using convolutional and mean pooling, corresponding to the enzyme-to-substrate, complex-level, and substrate-to-enzyme views, respectively. To adaptively balance the contributions of different interaction views, we
apply a channel-wise gating mechanism. Each view feature is first
transformed by a view-specific linear projection:
\begin{equation}
\tilde{\mathbf{f}}_e = \mathbf{f}_e\mathbf{W}_e,\quad
\tilde{\mathbf{f}}_c = \mathbf{f}_c\mathbf{W}_c,\quad
\tilde{\mathbf{f}}_s = \mathbf{f}_s\mathbf{W}_s,
\end{equation}
where $\{\mathbf{\tilde f}_e,\mathbf{\tilde f}_c,\mathbf{\tilde f}_s\}$ are the transformed features of the three interaction views. We then compute the gating logits and split them into three
view-specific vectors:
\begin{equation}
\mathbf{g}
=
\mathbf{W}_g[\tilde{\mathbf{f}}_e;\tilde{\mathbf{f}}_c;\tilde{\mathbf{f}}_s]
=
[\mathbf{g}_e;\mathbf{g}_c;\mathbf{g}_s],
\quad
\mathbf{g}_e,\mathbf{g}_c,\mathbf{g}_s \in \mathbb{R}^{d_a}.
\end{equation}
For each feature channel \(j\), the softmax is applied across the three
interaction views:
\begin{equation}
[\alpha_{e,j}, \alpha_{c,j}, \alpha_{s,j}]
= \rm softmax
\left(
[g_{e,j}, g_{c,j}, g_{s,j}]
\right),
\quad
j=1,\ldots,d_a.
\end{equation}
Finally, the fused feature $\mathbf{f}_F \in \mathbb{R}^{d_a}$ is computed
as
\begin{equation}
\mathbf{f}_F
=\boldsymbol{\alpha}_e\odot\tilde{\mathbf{f}}_e
+\boldsymbol{\alpha}_c\odot\tilde{\mathbf{f}}_c
+\boldsymbol{\alpha}_s\odot\tilde{\mathbf{f}}_s,
\end{equation}
where $\boldsymbol{\alpha}_v=[\alpha_{v,1},\ldots,\alpha_{v,d_a}]$ for $v\in\{e,c,s\}$ denote the channel-wise gating
weights for the three interaction views and $\odot$ denotes the element-wise product.

\subsubsection{Decoder Module} After obtaining the fused representation $\mathbf{f}_F$, we use a two-layer feed-forward decoder followed by a sigmoid function to predict enzyme-substrate reactivity:
\begin{equation}
P_m = \sigma(\mathbf{W}_2\phi(\mathbf{W}_1\mathbf{f}_F)),
\end{equation}
where $\phi(\cdot)$ denotes ${\rm ReLU}$ and $\sigma(\cdot)$ denotes the sigmoid function. The output $P_m \in (0,1)$ is the predicted probability that the given enzyme–substrate pair is reactive.

\subsection{The Retrieval Augmentation  Inference Strategy}
To improve robustness under low-sequence-identity evaluation, we introduce an inference-time retrieval augmentation strategy that uses neighborhood evidence from similar enzymes in representation space. As shown in Fig.~\ref{fig_frame}(b), the retrieval component is decoupled from model training and can be applied in a plug-and-play manner, avoiding the additional cost of retrieval-in-training pipelines. The strategy consists of three steps: embedding-based retrieval, retrieval signal aggregation, and prediction augmentation.

\subsubsection{Embedding-Based Retrieval} 
The first step retrieves representation-space neighbors for the query enzyme.

\paragraph{Offline Index Construction}
Let $\mathcal{E}_{\mathrm{train}}$ be the set of unique training enzymes. Each unique training enzyme $e_i \in \mathcal{E}_{\mathrm{train}}$ is encoded by a pretrained enzyme representation model into a retrieval representation $\mathbf{h}_i \in \mathbb{R}^{d_r}$, where $d_r$ is the retrieval hidden dimension. We construct a FAISS \cite{johnson2019billion} indexed retrieval memory bank $\mathcal{B}$, which stores the enzyme sequences and representations for efficient retrieval search.
\begin{equation}
\mathcal{B}=\{\langle e_i,\mathbf{h}_i\rangle\}_{e_i\in\mathcal{E}_{\mathrm{train}}}.
\end{equation}

\paragraph{Online Neighbor Retrieval}
During inference, for a query enzyme $e_q$, FAISS compares its retrieval representation with the indexed enzyme representations in $\mathcal{B}$ under a selected retrieval metric (\ie, cosine similarity), and returns the $K$ highest-ranked relevant enzymes, which form the retrieved neighbor set $\mathcal{N}_K(e_q)$:
\begin{equation}
\mathcal{N}_K(e_q)=\{e_1,e_2,...,e_i,...,e_K\}, e_i \in \mathcal{E}_{train}.
\end{equation}

\subsubsection{Retrieval Signal
Aggregation} Directly aggregating neighbor labels is insufficient for ESI prediction, 
because enzyme reactivity also depends on the query substrate. We therefore use model-based consensus aggregation. For each retrieved enzyme $e_n \in \mathcal{N}_K(e_q)$, we construct a support pair $(e_n,s_q)$ with the query substrate $s_q$ and use the trained RAMMESI model to obtain its interaction probability $\psi_\Theta(e_n,s_q)$. The retrieval signal is computed as a weighted consensus over the support predictions:
\begin{equation}
P_r =
\sum_{e_n \in \mathcal{N}_K(e_q)}
\beta_n \psi_\Theta(e_n,s_q),
\end{equation}
where $\beta_n$ denotes the normalized similarity-based weight.
\begin{equation}
\beta_n =
\frac{
\exp\left(\operatorname{sim}(\mathbf{h}_q,\mathbf{h}_n)\right)
}{
\sum_{e_j \in \mathcal{N}_K(e_q)}
\exp\left(\operatorname{sim}(\mathbf{h}_q,\mathbf{h}_j)\right)
},
\end{equation}
where $\mathbf{h}_q$, $\mathbf{h}_n$, and $\mathbf{h}_j$ are enzyme retrieval
representations, and $\operatorname{sim}(\cdot)$ denotes cosine similarity. The weights favor highly similar enzymes and down-weight distant neighbors.

\subsubsection{Prediction Augmentation}
Finally, we combine the base prediction $P_m$ of the query pair $(e_q,s_q)$ with the retrieval signal $P_r$ through linear interpolation. 
The final augmented probability $P_I$ is computed as:
\begin{equation}
P_I =(1-\lambda)\cdot P_m + \lambda\cdot P_r,
\end{equation}
where $\lambda \in [0, 1]$ controls the contribution of the retrieval signal. This interpolation injects neighborhood evidence into the final prediction and improves robustness under low-sequence-identity distribution shifts.

Algorithm~\ref{alg_inference} summarizes the proposed inference procedure. During inference, the additional computational overhead mainly comes from nearest-neighbor retrieval and extra model forward passes for the retrieved support pairs. FAISS indexing enables efficient similarity search over the memory bank, whose practical cost depends on the index type and memory-bank size. The retrieval-augmented signal requires $K$ additional forward passes, so the model-side inference cost scales linearly with the number of retrieved neighbors. As shown in Section~\ref{re_aug_sec}, a small value such as $K=5$ provides sufficient performance gains with limited latency overhead.

\begin{algorithm}[t]
\caption{Retrieval-Augmented Inference for RAMMESI}
\label{alg_inference}
\begin{algorithmic}[1]
\REQUIRE 
    Query pair $(e_q, s_q)$; 
    Trained RAMMESI model $\psi_\Theta$;
    Enzyme Memory Bank $\mathcal{B}=\{\langle e_i,\mathbf{h}_i\rangle\}_{e_i\in\mathcal{E}_{\mathrm{train}}}$;
    Hyperparameters: retrieval count $K$, weight $\lambda$.
\ENSURE Interaction Probability $P_I$.

\STATE \textbf{// Phase 1: Base Prediction}
\STATE Compute base probability: $P_m \leftarrow \psi_\Theta(e_q, s_q)$

\STATE \textbf{// Phase 2: Embedding-Based Retrieval}
\STATE Encode query enzyme to retrieval representation $\mathbf{h}_{q}$
\STATE Retrieve neighbor set $\mathcal{N}_K(e_q)$ from $\mathcal{B}$ 

\STATE \textbf{// Phase 3: Retrieval Signal Aggregation}
\IF{$\mathcal{N}_K(e_q) \neq \emptyset$}
    \STATE Initialize support evidence sum $S \leftarrow 0$
    \STATE Compute normalization denominator $D \leftarrow \sum_{e_j \in \mathcal{N}_K(e_q)} \exp(\text{sim}(\mathbf{h}_{q}, \mathbf{h}_{j}))$
    \FORALL{retrieved enzyme $e_n \in \mathcal{N}_K(e_q)$}
        \STATE Calculate importance weight: $\beta_n \leftarrow \frac{\exp(\text{sim}(\mathbf{h}_q, \mathbf{h}_n))}{D}$
        \STATE Predict support pair interaction: $P_{support} \leftarrow \psi_\Theta(e_n,s_q)$
        \STATE Accumulate weighted evidence: $S \leftarrow S + \beta_n \cdot P_{support}$
    \ENDFOR
    \STATE Set retrieval signal: $P_r \leftarrow S$
\ELSE
    \STATE $P_r \leftarrow P_m$ \quad \textit{// Fallback if no neighbors found}
\ENDIF

\STATE \textbf{// Phase 4: Prediction Augmentation}
\STATE Fuse probabilities: $P_I \leftarrow (1-\lambda)\cdot P_m + \lambda\cdot P_r$
\STATE \textbf{return} $P_I$
\end{algorithmic}
\end{algorithm}

\subsection{Optimization Objective}
For an enzyme-substrate pair with label $y \in \{0,1\}$, let $\hat{p} \in (0,1)$ denote the predicted probability for the reactive class. During training, $\hat{p}$ corresponds to the base prediction $P_m$. The binary cross-entropy (BCE) is frequently used as the default loss, which is
\begin{equation}
\mathcal{L}_{\mathrm{BCE}}(y,\hat{p}) = - y \log \hat{p} - (1-y)\log(1-\hat{p}).
\end{equation}

\subsubsection{Weighted BCE Framework for Severe Class Imbalance}
Although widely used for binary classification, BCE loss can be suboptimal under severe class imbalance. In ESI prediction, non-reactive pairs often outnumber reactive pairs \cite{cui2025enzyme}. This imbalance can make the training objective disproportionately influenced by easy negatives, which are non-reactive pairs ($y=0$) with low predicted probability $\hat{p}$. Consequently, the relative contribution of scarce positive examples may be weakened. We express imbalance-aware optimization under a weighted BCE formulation.
\begin{equation}
    \mathcal{L}(p_t) = - w(p_t)\log(p_t),
\end{equation}
where
\begin{equation}
    p_t = y\hat{p} + (1-y)(1-\hat{p})
\end{equation}
denotes the predicted probability assigned to the ground-truth class, with $p_t=\hat{p}$ for reactive pairs and $p_t=1-\hat{p}$ for non-reactive pairs. $p_t$ also serves as the target-class confidence, where a larger value indicates higher confidence in the ground-truth class \cite{lin2017focal}. Under this framework, different objectives correspond to different choices of the weighting function $w(p_t)$. In particular:
\begin{enumerate}
    \item \textit{Standard BCE loss} assigns a uniform weight to all samples:
    \begin{equation}
        w_{\mathrm{BCE}}(p_t)=1.
    \end{equation}
    \item \textit{Focal loss} introduces confidence-dependent power-law weights:
    \begin{equation}
        w_{\mathrm{FL}}(p_t)=(1-p_t)^\gamma, 
    \end{equation}
    where $\gamma$ controls the focusing strength. Larger $\gamma$ values down-weight high-confidence samples more strongly and place greater emphasis on samples with lower target class confidence. However, the focal loss weighting behavior is governed by a single parameter $\gamma$, which motivates a more flexible weighting variant.
    \item \textit{Threshold-aware smooth} (TAS) serves as a more flexible weighting variant:
    \begin{equation}
    \begin{split}
    w_{\mathrm{TAS}}(p_t) &= \rho \Bigl(\tanh\bigl(k((1-p_t)-\theta)\bigr) - \xi\Bigr).
    \end{split}
    \end{equation}
    Here, $\theta \in (0,1)$ primarily controls the transition location along the error axis $\epsilon=1-p_t$, while $k>0$ controls the transition sharpness around $\theta$. The constants $\rho$ and $\xi$ are deterministic functions of $k$ and $\theta$ used to normalize $w_{\mathrm{TAS}}$ to $[0,1]$.
\end{enumerate}

\begin{figure*}[!t]
\centering
{\includegraphics[width=\linewidth]{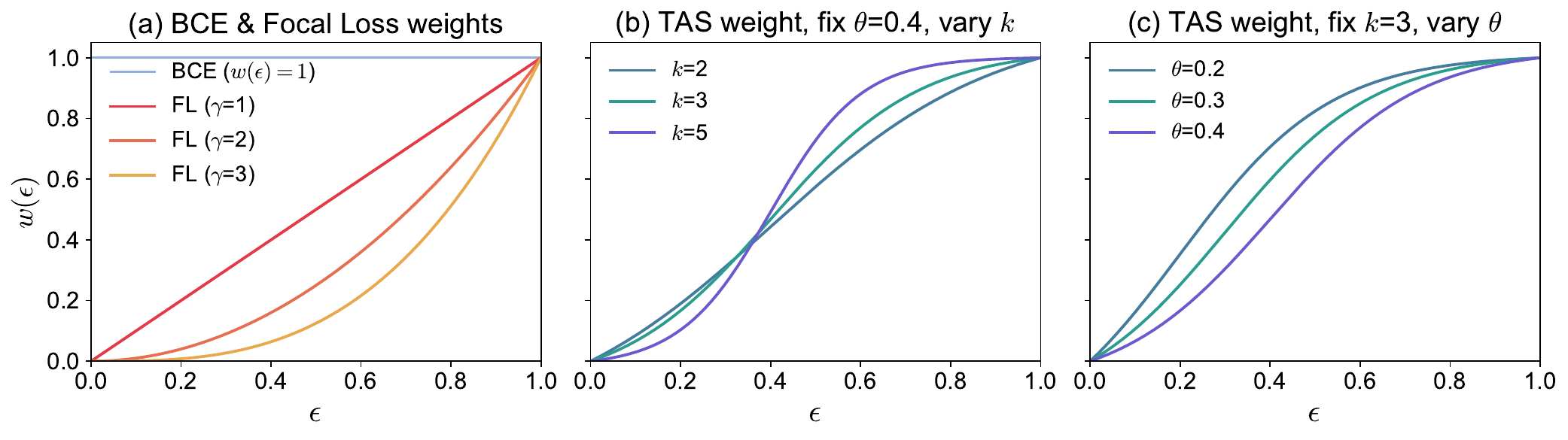}}
\caption{Comparison of weighting functions under the weighted-BCE framework. Here, $w(\epsilon)$ denotes the sample weight, $\epsilon=1-p_t$ denotes the prediction error. (a) BCE uses a constant weight, while focal loss uses power-law modulation in which $\gamma$ jointly changes the focusing region and curve shape. (b) In TAS, varying $k$ mainly adjusts transition sharpness. (c) In TAS, varying $\theta$ mainly shifts the transition location.}
\label{fig_loss_we}
\end{figure*}

Under this unified formulation, BCE, focal loss, and TAS can be interpreted as different choices of the sample-weighting function. This view provides a clearer basis for loss design when prior knowledge about label imbalance or sample difficulty is available. Compared with the uniform weighting in BCE, focal loss and TAS assign confidence-dependent weights to training samples, thereby reducing the dominance of easy negative examples under severe ESI class imbalance.

The key difference between focal loss and TAS lies in how the weighting behavior is parameterized. In focal loss, the single parameter $\gamma$ jointly controls the focusing region and the curve steepness. In contrast, TAS separates these two effects through $\theta$ and $k$. As illustrated in Fig.~\ref{fig_loss_we}, $\theta$ primarily shifts the transition location, whereas $k$ primarily adjusts the transition sharpness. This design enables more flexible sample reweighting when the desired focusing region varies across datasets or evaluation settings.

\section{Experiments}
\label{exp}

We evaluate RAMMESI with respect to the three main challenges identified above: substrate-conditioned pairwise modeling, low-homology generalization, and learning under class imbalance. Specifically, we compare RAMMESI with existing ESI predictors on two public datasets under sequence-identity-aware evaluation, analyze its behavior across different imbalance settings, and conduct ablations to quantify the contribution of key components. We further examine the effect of inference-time retrieval augmentation and present a case study on enzyme-substrate candidate prioritization.

\subsection{Experimental Protocols}
\paragraph{Benchmarks}
All experiments are conducted on two public ESI benchmarks: \textbf{ESP-DB}~\cite{kroll2023general} and \textbf{Reactzyme-DB}~\cite{hua2024reactzyme}. ESP-DB is released as enzyme-substrate pairs with binary interaction labels. In ESP-DB, negative pairs are constructed by sampling structurally similar molecules that are not annotated as substrates of the corresponding enzyme, followed by an enzyme-sequence-identity-aware split between the training and test sets. Reactzyme-DB is originally curated at the reaction level, where each enzyme is associated with a catalyzed reaction equation rather than explicit enzyme-substrate pairs. To adapt Reactzyme-DB to pairwise ESI prediction and ensure consistency with ESP-DB, we convert each reaction into enzyme-substrate positive pairs using molecular reactants as candidate substrates. After converting Reactzyme-DB into pair-level positive pairs, we generate negative pairs using the similarity-based sampling strategy adopted in ESP-DB and follow the same splitting protocol to construct training, validation, and test sets. For preprocessing, we remove ESP-DB pairs whose substrates cannot be embedded by the molecular encoder and exclude non-standard Reactzyme-DB participants such as water, gases, and metal ions. Table~\ref{tab_dataset_stats} summarizes the processed datasets. Furthermore, to evaluate generalization beyond close homologs, we compute the maximum sequence identity between each test enzyme and the training enzymes using MMseqs2 \cite{steinegger2017mmseqs2}. Results are reported in three low-identity intervals of $<20\%$, $[20\%,30\%)$, and $[30\%,40\%)$, where lower identity indicates a more challenging evaluation
setting. Detailed statistics of the three test sets across different identity intervals are reported in Table S3 in the Appendix.

\begin{table}[t]
\centering
\caption{Dataset statistics.}
\label{tab_dataset_stats}
\resizebox{\columnwidth}{!}{%
\begin{tabular}{lcccc}
\toprule
\textbf{Dataset} & \textbf{\# Enzymes} & \textbf{\# Substrates} & \textbf{\# Total Pairs} & \textbf{Pos.:Neg.} \\
\midrule
ESP-DB & 146,613 & 1,336 & 782,273   & 1:2.7 \\
Reactzyme-DB & 125,092 & 5,529 & 657,495 & 1:2.6 \\
\bottomrule
\end{tabular}
}
\end{table}

\paragraph{Implementation}

RAMMESI is trained with AdamW \cite{loshchilov2018decoupled} using a learning rate of $3 \times 10^{-5}$, a batch size of $256$, and early stopping based on validation AUROC for up to $50$ epochs. The attention hidden dimension $d_a$ is set to $768$, and the joint modeling transformer contains $3$ layers. For Threshold-Aware Smooth weighting, we set $k=3,\theta=0.4$ on ESP-DB and $k=2,\theta=0.4$ on Reactzyme-DB.

\paragraph{Baselines}
We compare RAMMESI with traditional machine learning baselines, a simple neural baseline, and existing ESI prediction deep learning methods. RandomForest~\cite{breiman2001random}, LightGBM~\cite{ke2017lightgbm}, and DNN are implemented using concatenated mean-pooled representations from ESM2-650M~\cite{lin2023evolutionary} for enzymes and UniMol2-570M~\cite{ji2024uni} for substrates. 
We further compare with ProSmith~\cite{kroll2024multimodal}, VIPER~\cite{campbell2024viper}, ESP~\cite{kroll2023general}, OmniESI~\cite{nie2025omniesi}, and FusionESP~\cite{du2025fusionesp}, which represent existing deep learning approaches for ESI prediction. 
For published baselines, we use official implementations with default hyperparameter settings; implementation sources are summarized in Table S4 in the Appendix.

\paragraph{Evaluation}
We formulate ESI prediction as a binary classification task, where each enzyme-substrate pair is labeled as reactive (positive) or non-reactive (negative). To account for label imbalance, we evaluate model performance using Recall, AUROC, AUPRC, and F1 score. In particular, AUROC and AUPRC measure ranking quality, while F1 score and Recall assess positive-class performance.

\begin{table*}[htbp]
\centering
\caption{Performance Comparison on ESP-DB and Reactzyme-DB.}
\label{tab_exp_result}

\setlength{\tabcolsep}{3.2pt} 
\resizebox{\textwidth}{!}{%
\begin{tabular}{c|l|cccc|cccc}
\toprule
\multirow{2}{*}{\textbf{Identity}} & \multirow{2}{*}{\textbf{Model}} & \multicolumn{4}{c|}{\textbf{ESP-DB}} & \multicolumn{4}{c}{\textbf{Reactzyme-DB}} \\
\cline{3-10}
& & \rule{0pt}{2.6ex}Recall & AUROC & AUPRC & F1 & Recall & AUROC & AUPRC & F1 \\
\midrule
\multirow{9}{*}{\textless 20\%} 
& RandomForest & 0.091{\scriptsize$\pm$0.014} & 0.717{\scriptsize$\pm$0.015} & 0.522{\scriptsize$\pm$0.023} & 0.165{\scriptsize$\pm$0.023} & 0.178{\scriptsize$\pm$0.018} & 0.733{\scriptsize$\pm$0.015} & 0.611{\scriptsize$\pm$0.022} & 0.301{\scriptsize$\pm$0.025} \\
& LightGBM     & 0.033{\scriptsize$\pm$0.009} & 0.665{\scriptsize$\pm$0.016} & 0.463{\scriptsize$\pm$0.023} & 0.064{\scriptsize$\pm$0.018} & 0.037{\scriptsize$\pm$0.008} & 0.613{\scriptsize$\pm$0.016} & 0.443{\scriptsize$\pm$0.023} & 0.073{\scriptsize$\pm$0.017} \\
& DNN          & \underline{0.512{\scriptsize$\pm$0.023}} & 0.824{\scriptsize$\pm$0.011} & 0.695{\scriptsize$\pm$0.018} & 0.595{\scriptsize$\pm$0.019} & 0.440{\scriptsize$\pm$0.024} & 0.767{\scriptsize$\pm$0.013} & 0.616{\scriptsize$\pm$0.022} & 0.515{\scriptsize$\pm$0.021} \\
& ProSmith     & 0.336{\scriptsize$\pm$0.021} & 0.588{\scriptsize$\pm$0.013} & 0.338{\scriptsize$\pm$0.017} & 0.360{\scriptsize$\pm$0.019} & 0.187{\scriptsize$\pm$0.022} & 0.620{\scriptsize$\pm$0.017} & 0.417{\scriptsize$\pm$0.026} & 0.282{\scriptsize$\pm$0.028} \\
& VIPER        & 0.015{\scriptsize$\pm$0.007} & 0.523{\scriptsize$\pm$0.017} & 0.272{\scriptsize$\pm$0.017} & 0.028{\scriptsize$\pm$0.012} & 0.042{\scriptsize$\pm$0.010} & 0.486{\scriptsize$\pm$0.016} & 0.285{\scriptsize$\pm$0.016} & 0.075{\scriptsize$\pm$0.016} \\
& ESP          & 0.485{\scriptsize$\pm$0.023} & 0.807{\scriptsize$\pm$0.011} & 0.659{\scriptsize$\pm$0.018} & 0.545{\scriptsize$\pm$0.021} & 0.391{\scriptsize$\pm$0.033} & 0.758{\scriptsize$\pm$0.021} & 0.590{\scriptsize$\pm$0.036} & 0.487{\scriptsize$\pm$0.033} \\
& OmniESI      & 0.503{\scriptsize$\pm$0.024} & 0.828{\scriptsize$\pm$0.013} & 0.709{\scriptsize$\pm$0.020} & 0.600{\scriptsize$\pm$0.022} & \underline{0.481{\scriptsize$\pm$0.026}} & \underline{0.795{\scriptsize$\pm$0.012}} & \underline{0.685{\scriptsize$\pm$0.020}} & \underline{0.545{\scriptsize$\pm$0.023}} \\
& FusionESP    & 0.467{\scriptsize$\pm$0.004} & \underline{0.844{\scriptsize$\pm$0.007}} & \textbf{0.758{\scriptsize$\pm$0.003}} & \underline{0.608{\scriptsize$\pm$0.001}} & 0.376{\scriptsize$\pm$0.010} & 0.752{\scriptsize$\pm$0.004} & 0.670{\scriptsize$\pm$0.001} & 0.517{\scriptsize$\pm$0.011} \\
& RAMMESI & \textbf{0.553{\scriptsize$\pm$0.024}} & \textbf{0.848{\scriptsize$\pm$0.002}} & \underline{0.736{\scriptsize$\pm$0.019}} & \textbf{0.616{\scriptsize$\pm$0.021}} & \textbf{0.487{\scriptsize$\pm$0.023}} & \textbf{0.833{\scriptsize$\pm$0.011}} & \textbf{0.728{\scriptsize$\pm$0.018}} & \textbf{0.578{\scriptsize$\pm$0.021}} \\
\midrule
\multirow{9}{*}{[20\%, 30\%)} 
& RandomForest & 0.190{\scriptsize$\pm$0.037} & 0.818{\scriptsize$\pm$0.023} & 0.693{\scriptsize$\pm$0.041} & 0.315{\scriptsize$\pm$0.052} & 0.187{\scriptsize$\pm$0.025} & 0.778{\scriptsize$\pm$0.017} & 0.641{\scriptsize$\pm$0.026} & 0.310{\scriptsize$\pm$0.037} \\
& LightGBM     & 0.061{\scriptsize$\pm$0.021} & 0.621{\scriptsize$\pm$0.031} & 0.420{\scriptsize$\pm$0.045} & 0.112{\scriptsize$\pm$0.037} & 0.015{\scriptsize$\pm$0.006} & 0.606{\scriptsize$\pm$0.021} & 0.449{\scriptsize$\pm$0.030} & 0.011{\scriptsize$\pm$0.011} \\
& DNN          & 0.612{\scriptsize$\pm$0.046} & 0.872{\scriptsize$\pm$0.018} & 0.774{\scriptsize$\pm$0.033} & 0.663{\scriptsize$\pm$0.038} & 0.488{\scriptsize$\pm$0.036} & 0.779{\scriptsize$\pm$0.019} & 0.659{\scriptsize$\pm$0.031} & 0.545{\scriptsize$\pm$0.032} \\
& ProSmith     & 0.358{\scriptsize$\pm$0.041} & 0.601{\scriptsize$\pm$0.033} & 0.362{\scriptsize$\pm$0.035} & 0.377{\scriptsize$\pm$0.038} & 0.211{\scriptsize$\pm$0.030} & 0.593{\scriptsize$\pm$0.027} & 0.425{\scriptsize$\pm$0.033} & 0.320{\scriptsize$\pm$0.038} \\
& VIPER        & 0.037{\scriptsize$\pm$0.018} & 0.503{\scriptsize$\pm$0.030} & 0.284{\scriptsize$\pm$0.032} & 0.066{\scriptsize$\pm$0.032} & 0.063{\scriptsize$\pm$0.019} & 0.497{\scriptsize$\pm$0.024} & 0.288{\scriptsize$\pm$0.025} & 0.106{\scriptsize$\pm$0.030} \\
& ESP          & 0.625{\scriptsize$\pm$0.048} & 0.886{\scriptsize$\pm$0.019} & 0.801{\scriptsize$\pm$0.031} & 0.691{\scriptsize$\pm$0.038} & 0.343{\scriptsize$\pm$0.048} & 0.682{\scriptsize$\pm$0.030} & 0.483{\scriptsize$\pm$0.048} & 0.414{\scriptsize$\pm$0.050} \\
& OmniESI      & \underline{0.682{\scriptsize$\pm$0.045}} & \underline{0.915{\scriptsize$\pm$0.015}} & \underline{0.854{\scriptsize$\pm$0.023}} & \underline{0.743{\scriptsize$\pm$0.031}} & \underline{0.566{\scriptsize$\pm$0.036}} & \underline{0.834{\scriptsize$\pm$0.016}} & 0.728{\scriptsize$\pm$0.027} & \underline{0.604{\scriptsize$\pm$0.030}} \\
& FusionESP    & 0.626{\scriptsize$\pm$0.012} & 0.891{\scriptsize$\pm$0.018} & 0.842{\scriptsize$\pm$0.016} & 0.724{\scriptsize$\pm$0.014} & 0.457{\scriptsize$\pm$0.003} & 0.817{\scriptsize$\pm$0.014} & \underline{0.735{\scriptsize$\pm$0.004}} & 0.578{\scriptsize$\pm$0.005} \\
& RAMMESI & \textbf{0.728{\scriptsize$\pm$0.036}} & \textbf{0.942{\scriptsize$\pm$0.011}} & \textbf{0.883{\scriptsize$\pm$0.020}} & \textbf{0.783{\scriptsize$\pm$0.028}} & \textbf{0.583{\scriptsize$\pm$0.030}} & \textbf{0.842{\scriptsize$\pm$0.017}} & \textbf{0.745{\scriptsize$\pm$0.024}} & \textbf{0.628{\scriptsize$\pm$0.026}} \\
\midrule
\multirow{9}{*}{[30\%, 40\%)} 
& RandomForest & 0.250{\scriptsize$\pm$0.027} & 0.835{\scriptsize$\pm$0.015} & 0.708{\scriptsize$\pm$0.024} & 0.399{\scriptsize$\pm$0.034} & 0.282{\scriptsize$\pm$0.014} & 0.847{\scriptsize$\pm$0.009} & 0.738{\scriptsize$\pm$0.013} & 0.433{\scriptsize$\pm$0.017} \\
& LightGBM     & 0.122{\scriptsize$\pm$0.018} & 0.739{\scriptsize$\pm$0.015} & 0.574{\scriptsize$\pm$0.027} & 0.216{\scriptsize$\pm$0.028} & 0.029{\scriptsize$\pm$0.006} & 0.674{\scriptsize$\pm$0.010} & 0.477{\scriptsize$\pm$0.016} & 0.056{\scriptsize$\pm$0.010} \\
& DNN          & 0.696{\scriptsize$\pm$0.027} & 0.906{\scriptsize$\pm$0.010} & 0.815{\scriptsize$\pm$0.019} & 0.718{\scriptsize$\pm$0.020} & 0.526{\scriptsize$\pm$0.015} & 0.874{\scriptsize$\pm$0.006} & 0.754{\scriptsize$\pm$0.012} & 0.603{\scriptsize$\pm$0.013} \\
& ProSmith     & 0.361{\scriptsize$\pm$0.026} & 0.593{\scriptsize$\pm$0.021} & 0.360{\scriptsize$\pm$0.024} & 0.382{\scriptsize$\pm$0.025} & 0.218{\scriptsize$\pm$0.016} & 0.633{\scriptsize$\pm$0.012} & 0.458{\scriptsize$\pm$0.018} & 0.321{\scriptsize$\pm$0.019} \\
& VIPER        & 0.025{\scriptsize$\pm$0.010} & 0.486{\scriptsize$\pm$0.020} & 0.262{\scriptsize$\pm$0.021} & 0.047{\scriptsize$\pm$0.017} & 0.032{\scriptsize$\pm$0.006} & 0.474{\scriptsize$\pm$0.012} & 0.277{\scriptsize$\pm$0.012} & 0.057{\scriptsize$\pm$0.010} \\
& ESP          & 0.708{\scriptsize$\pm$0.030} & 0.909{\scriptsize$\pm$0.012} & 0.831{\scriptsize$\pm$0.018} & 0.738{\scriptsize$\pm$0.021} & 0.547{\scriptsize$\pm$0.019} & 0.823{\scriptsize$\pm$0.010} & 0.732{\scriptsize$\pm$0.015} & 0.622{\scriptsize$\pm$0.015} \\
& OmniESI      & \underline{0.766{\scriptsize$\pm$0.028}} & \underline{0.942{\scriptsize$\pm$0.009}} & \underline{0.896{\scriptsize$\pm$0.015}} & \underline{0.803{\scriptsize$\pm$0.019}} & \textbf{0.616{\scriptsize$\pm$0.015}} & \underline{0.873{\scriptsize$\pm$0.006}} & 0.790{\scriptsize$\pm$0.011} & \underline{0.681{\scriptsize$\pm$0.012}} \\
& FusionESP    & 0.730{\scriptsize$\pm$0.002} & 0.924{\scriptsize$\pm$0.002} & 0.873{\scriptsize$\pm$0.003} & 0.797{\scriptsize$\pm$0.006} & 0.566{\scriptsize$\pm$0.013} & 0.867{\scriptsize$\pm$0.005} & \textbf{0.804{\scriptsize$\pm$0.007}} & 0.680{\scriptsize$\pm$0.009} \\
& RAMMESI & \textbf{0.846{\scriptsize$\pm$0.023}} & \textbf{0.951{\scriptsize$\pm$0.011}} & \textbf{0.908{\scriptsize$\pm$0.014}} & \textbf{0.854{\scriptsize$\pm$0.016}} & \underline{0.614{\scriptsize$\pm$0.016}} & \textbf{0.876{\scriptsize$\pm$0.006}} & \underline{0.803{\scriptsize$\pm$0.010}} & \textbf{0.686{\scriptsize$\pm$0.013}} \\
\bottomrule
\end{tabular}}
\\[2pt]
\vspace{2pt}
\parbox{\textwidth}{\footnotesize\raggedright
\hspace*{1em}Optimal results are in \textbf{bold}; second-best results are \underline{underlined}.
}
\end{table*}

\subsection{Baseline Comparison}
Table~\ref{tab_exp_result} compares RAMMESI with baseline methods on ESP-DB and Reactzyme-DB under low-sequence-identity evaluation. Results are reported for three identity regimes, \ie, $<20\%$, $[20\%,30\%)$, and $[30\%,40\%)$, to assess performance under low-homology shift. As sequence identity between test enzymes and training enzymes decreases, most methods show clear performance degradation, confirming the difficulty of low-identity ESI prediction. Nevertheless, RAMMESI remains robust across identity regimes and achieves leading performance in most settings on both datasets. In particular, it obtains strong AUROC and Recall on ESP-DB and consistently strong performance on Reactzyme-DB, indicating better retention and ranking of reactive pairs when test enzymes have limited homology to the training set. FusionESP and OmniESI also perform competitively, further supporting the value of explicit cross-modal interaction modeling for ESI prediction.

We further assess robustness under class imbalance by comparing RAMMESI with top-performing baselines at positive-to-negative ratios from $1:5$ to $1:50$. Here we focus on AUPRC and Recall (reported in Fig.~\ref{fig_ratio_rank} AUROC and F1 results are provided in Fig.~S1 in the Appendix. AUPRC reflects ranking quality under imbalance, whereas Recall measures how many reactive pairs are retained for downstream screening. For Recall, RAMMESI consistently achieves the best performance across imbalance ratios on both datasets, indicating stronger retention of reactive enzyme-substrate pairs for downstream screening. For AUPRC, all methods degrade as the proportion of negative samples increases, suggesting that ranking positives above a larger pool of negatives becomes increasingly difficult. Nevertheless, RAMMESI maintains the highest AUPRC in most settings, indicating stronger ranking quality under severe imbalance. Together, its stable Recall and AUPRC show that RAMMESI better preserves positive candidates while maintaining their ranking, whereas baseline methods tend to lose either positive-class coverage or ranking precision under heavier imbalance.

\begin{figure}[!t]
\centering
{\includegraphics[width=\linewidth]{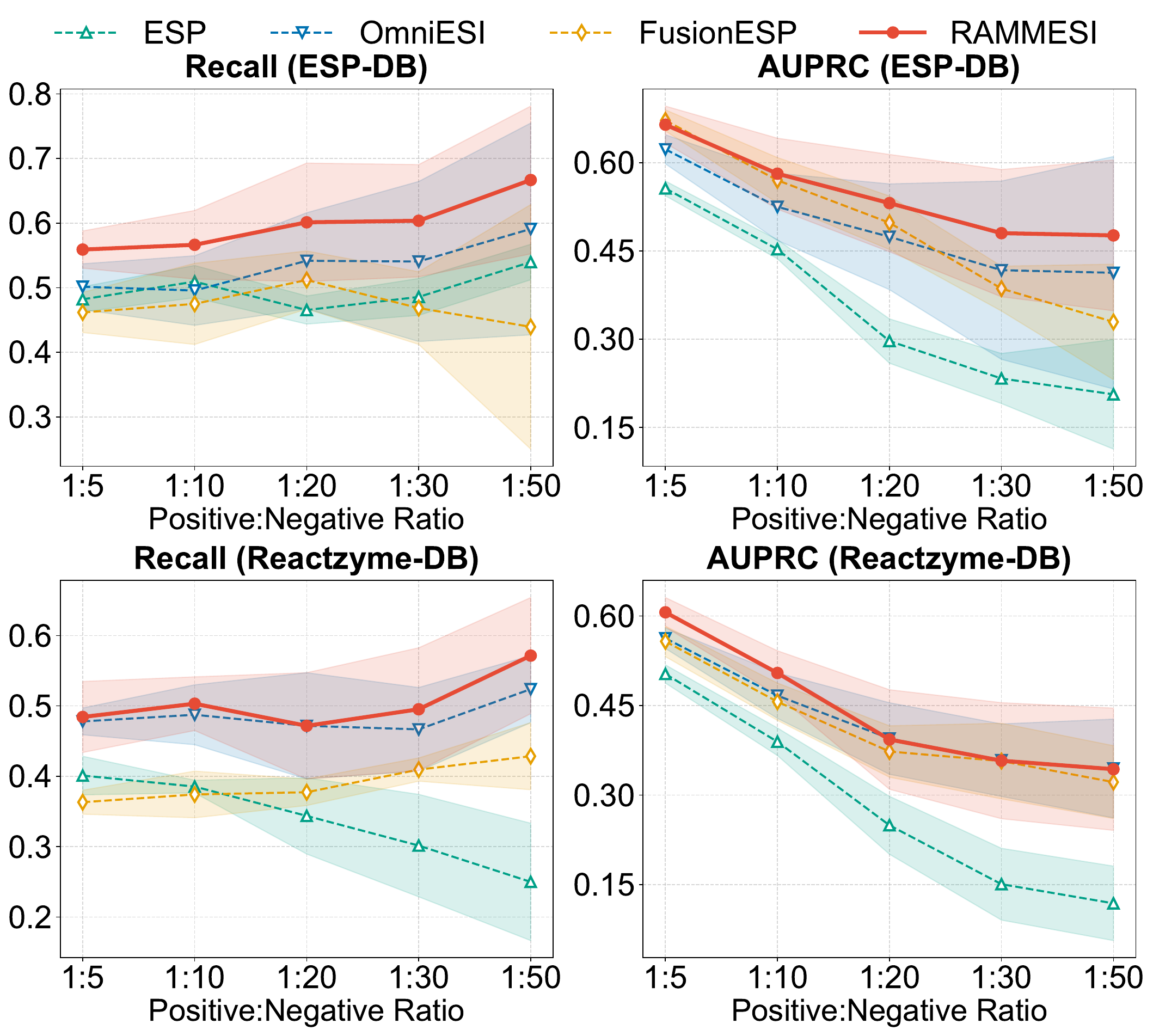}}
\caption{Recall and AUPRC performance on ESP-DB and Reactzyme-DB with different positive-to-negative ratios.}
\label{fig_ratio_rank}
\end{figure}

\begin{table}[t]
\centering
\caption{Model component ablation results under $<20\%$ identity.}
\label{tab_ab_res}
\resizebox{\columnwidth}{!}{%
\begin{tabular}{l l cccc}
\toprule
\textbf{Dataset} & \textbf{Setting} & \textbf{Recall} & \textbf{AUROC} & \textbf{AUPRC} & \textbf{F1} \\
\midrule
\multirow{4}{*}{ESP-DB}
& RAMMESI             & \textbf{0.555} & \textbf{0.848} & \textbf{0.737} & 0.618 \\
& w/o Gate            & 0.550 & 0.832 & 0.722  & 0.609 \\
& w/o Joint           & \textbf{0.555} & 0.824 & 0.731          & \textbf{0.639} \\
& w/o Joint $\&$ Gate & 0.547 & 0.822        & 0.710 & 0.611 \\
\midrule
\multirow{4}{*}{Reactzyme-DB} 
& RAMMESI             & 0.485 & \textbf{0.832} & \textbf{0.726} & \textbf{0.578} \\
& w/o Gate            & \textbf{0.498} & 0.777 & 0.671          & 0.572 \\
& w/o Joint           & 0.452 & 0.775        & 0.651 & 0.525 \\
& w/o Joint $\&$ Gate & 0.485 & 0.787        & 0.674 & 0.572 \\
\bottomrule
\end{tabular}
}
\end{table}

\begin{table}[ht]
\centering
\caption{Performance comparison of different loss functions under the $<20\%$ identity setting, with and without retrieval augmentation.}
\label{tab_loss_func}
\resizebox{\columnwidth}{!}{%
\begin{tabular}{llccccc}
\toprule
\textbf{Dataset} & \textbf{Loss} & \textbf{Retrieval} & \textbf{Recall} & \textbf{AUROC} & \textbf{AUPRC} & \textbf{F1} \\
\midrule
\multirow{6}{*}{ESP-DB}
& BCE   & \multirow{3}{*}{\centering \xmark} & 0.498 & 0.822 & 0.699 & 0.584 \\
& Focal &                                      & \textbf{0.578} & \textbf{0.843} & \textbf{0.740} & \textbf{0.641} \\
& TAS   &                                      & 0.568 & 0.829 & 0.715 & 0.614 \\
\cmidrule(lr){2-7}
& BCE   & \multirow{3}{*}{\centering \cmark} & 0.495 & 0.850 & 0.734 & 0.588 \\
& Focal &                                      & \textbf{0.573} & \textbf{0.855} & \textbf{0.755} & \textbf{0.658} \\
& TAS   &                                      & 0.555 & 0.848 & 0.737 & 0.618 \\
\midrule
\multirow{6}{*}{Reactzyme-DB}
& BCE   & \multirow{3}{*}{\centering \xmark} & 0.478 & 0.764 & 0.659 & 0.568 \\
& Focal &                                      & 0.473 & 0.767 & 0.653 & 0.550 \\
& TAS   &                                      & \textbf{0.491} & \textbf{0.781} & \textbf{0.678} & \textbf{0.576} \\
\cmidrule(lr){2-7}
& BCE   & \multirow{3}{*}{\centering \cmark} & 0.478 & 0.820 & \textbf{0.731} & 0.571 \\
& Focal &                                      & 0.464 & 0.802 & 0.695 & 0.559 \\
& TAS   &                                      & \textbf{0.485} & \textbf{0.832} & 0.726 & \textbf{0.578} \\
\bottomrule
\end{tabular}
}
\end{table}

\subsection{Ablation Study}
To validate key RAMMESI components, we conduct a set of ablation studies under the most challenging $<20\%$ sequence-identity setting. This subsection examines three aspects of our approach. (1) Model component ablations: we remove the key architectural modules to assess their contribution to pairwise enzyme–substrate interaction modeling. (2) Loss function comparison: we compare alternative loss functions (\eg, BCE loss and focal loss) to quantify the impact of imbalance-aware optimization under skewed ESI supervision. (3) Effect of the temperature factor: We report and discuss the performance impact of enabling the temperature factor.

\subsubsection{Model component ablations} 
We first compare the full model with three variants: removing the adaptive gating mechanism (w/o Gate), removing the joint modeling branch (w/o Joint), and removing both components (w/o Joint \& Gate).

As shown in Table~\ref{tab_ab_res}, the full model achieves the best AUROC and AUPRC on both datasets, indicating that joint interaction modeling and adaptive gating fusion improve discriminative performance under low-sequence-identity evaluation. Removing either component generally degrades performance, while removing both leads to a further decline, suggesting that the two modules provide complementary benefits. The joint branch contributes complex-level interaction context, whereas the gating mechanism adaptively integrates multiple interaction views for prediction.

The two datasets show different sensitivity patterns. On ESP-DB, RAMMESI obtains the best AUROC ($0.848$) and AUPRC ($0.737$), while the w/o Joint variant achieves a slightly higher F1, suggesting that the cross-attention branches alone already capture strong pairwise features on this dataset. On Reactzyme-DB, removing the joint branch causes the largest drops in AUROC, AUPRC, and F1, indicating a stronger need for complex-level context. The w/o Gate variant obtains slightly higher Recall, but with clear AUROC and AUPRC degradation, suggesting reduced ranking quality despite higher positive-class coverage. Overall, the ablation results show that the proposed component design improves RAMMESI for low-identity ESI prediction. Results for the other identity intervals are reported in Table~S5 in the Appendix.

\subsubsection{Loss function comparison} 
We compare BCE, focal loss, and TAS to examine how different weighted-BCE objectives perform across two datasets. Each loss is evaluated with and without retrieval augmentation. As shown in Table~\ref{tab_loss_func}, both focal loss and TAS generally improve over BCE, indicating that confidence-aware sample weighting is beneficial under skewed ESI supervision.

The relative advantage of focal loss and TAS, however, differs across datasets. On ESP-DB, focal loss achieves the strongest overall performance, suggesting that its power-law modulation is well matched to this dataset. On Reactzyme-DB, TAS shows more stable behavior across standard and retrieval-augmented settings: while focal loss does not consistently improve over BCE and underperforms on several metrics, TAS maintains gains in AUROC and F1 in both settings. These results indicate that TAS should not be viewed as a universal replacement for focal loss; rather, it provides a more flexible weighting variant when the optimal focusing profile differs across datasets. Results on other identity intervals are reported in Table~S6 in the Appendix.

\begin{table}[ht]
\centering
\caption{Performance comparison of temperature factor impact under the $<20\%$ identity setting, with and without retrieval augmentation.}
\label{tab_temp_res}
\resizebox{\columnwidth}{!}{%
\begin{tabular}{llccccc}
\toprule
\textbf{Dataset} & \textbf{Setting} & \textbf{Retrieval} & \textbf{Recall} & \textbf{AUROC} & \textbf{AUPRC} & \textbf{F1} \\
\midrule
\multirow{4}{*}{ESP-DB}
& Temp. Off & \multirow{2}{*}{\makebox[1.5em][c]{\xmark}} & 0.565 & 0.819 & \textbf{0.715} & \textbf{0.626} \\
& Temp. On  &  
& \textbf{0.568} 
& \textbf{0.829} 
& \textbf{0.715} & 0.614 \\
\cmidrule(lr){2-7}
& Temp. Off & \multirow{2}{*}{\makebox[1.5em][c]{\cmark}} & 0.550 & 0.843 & \textbf{0.742} & \textbf{0.625} \\
& Temp. On  &          & \textbf{0.555} 
& \textbf{0.848} 
& 0.737 & 0.618 \\
\midrule
\multirow{4}{*}{Reactzyme-DB}
& Temp. Off & \multirow{2}{*}{\makebox[1.5em][c]{\xmark}} & 0.466 & 0.775 & 0.659 & 0.550 \\
& Temp. On  &           & \textbf{0.491} 
& \textbf{0.781} 
& \textbf{0.678} 
& \textbf{0.576} \\
\cmidrule(lr){2-7}
& Temp. Off & \multirow{2}{*}{\makebox[1.5em][c]{\cmark}} & 0.468 & 0.817 & 0.710 & 0.563 \\
& Temp. On  &          & \textbf{0.485} 
& \textbf{0.832} 
& \textbf{0.726} 
& \textbf{0.578} \\
\bottomrule
\end{tabular}
}
\end{table}

\subsubsection{Effect of the temperature factor} 
Table~\ref{tab_temp_res} evaluates the temperature factor $\tau$ used to calibrate cross-modal attention scores in the joint interaction branch. 

Across both datasets, enabling temperature scaling improves AUROC and Recall in both standard and retrieval-augmented settings, suggesting that calibrated cross-modal attention is beneficial for low-identity ESI prediction. The effect is more pronounced on Reactzyme-DB, where temperature scaling improves all four metrics, including Recall from $0.466$ to $0.491$ and AUPRC from $0.659$ to $0.678$. On ESP-DB, temperature scaling improves AUROC and Recall, but slightly decreases AUPRC and F1, indicating a trade-off between positive-class retention and ranking precision. Overall, these results suggest that the temperature factor improves robustness-oriented metrics, especially Recall, although its effect varies across datasets and evaluation metrics. Results for the remaining identity intervals are reported in Table~S7.

\subsection{Retrieval Augmentation Analysis}
\label{re_aug_sec}
\begin{figure*}[!t]
\centering
{\includegraphics[width=\linewidth]{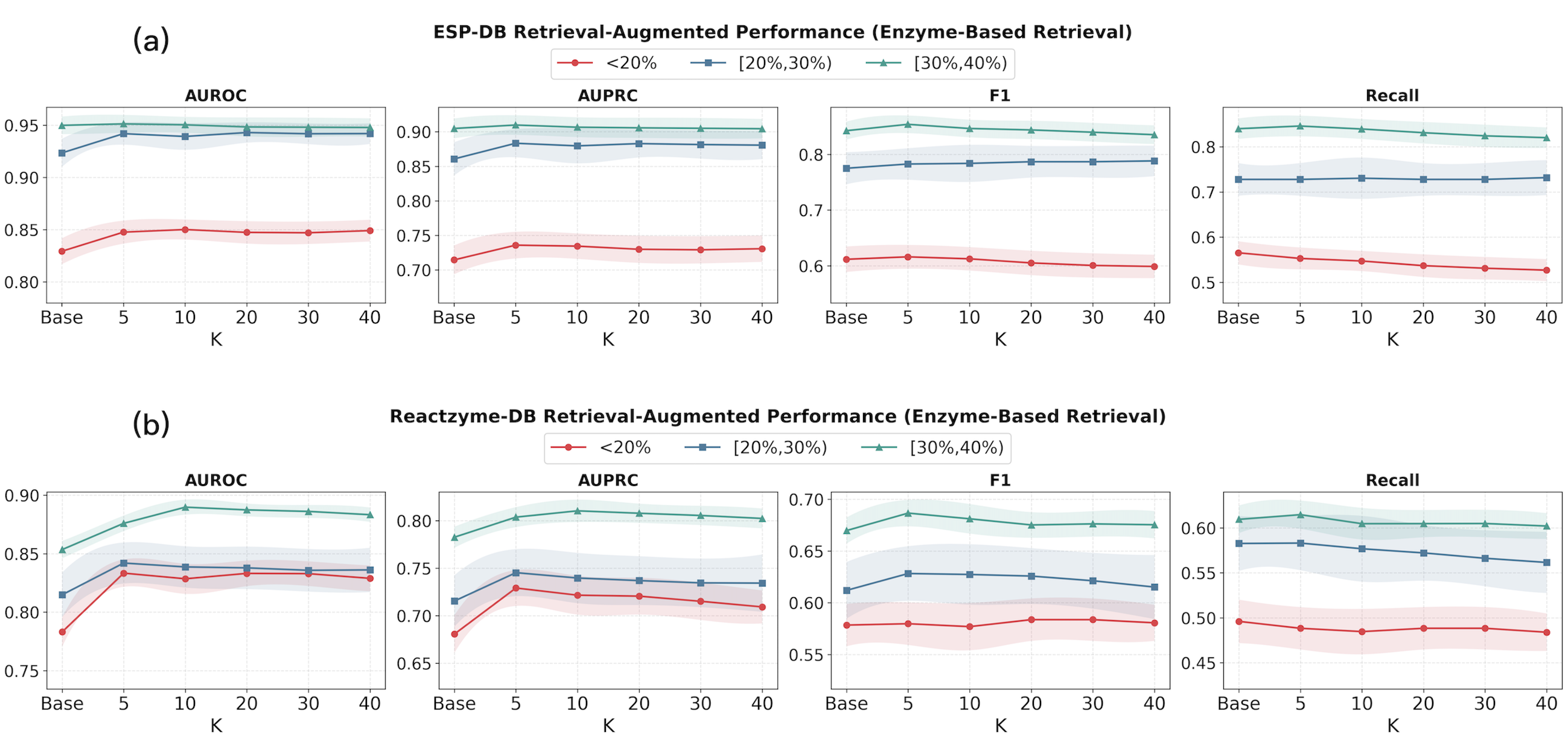}}
\caption{Inference-time retrieval performance on two datasets with different numbers of neighbors. Base is the model without retrieval augmentation, K is the number of retrieved neighbors. (a) The inference-time retrieval-augmentation performance on ESP-DB. (b) The inference-time retrieval-augmentation performance on Reactzyme-DB.}
\label{fig_re_k}
\end{figure*}

\begin{table*}[t]
\centering
\caption{Performance comparison with Plug-and-play retrieval augmentation on two ESI baselines ($K{=}5$).}
\label{tab_re_plug}
\begin{tabular}{l c cccc cccc}
\toprule
\multirow{2}{*}{\textbf{Identity}} & \multirow{2}{*}{\textbf{Retrieval}} &
\multicolumn{4}{c}{\textbf{ESP-DB}} & \multicolumn{4}{c}{\textbf{Reactzyme-DB}} \\
\cmidrule(lr){3-6}\cmidrule(lr){7-10}
& & Recall & AUROC & AUPRC & F1
& Recall & AUROC & AUPRC & F1 \\
\midrule

\multicolumn{10}{c}{\textbf{OmniESI}} \\
\midrule
\multirow{2}{*}{$<$ 20\%}
& \xmark & \textbf{0.505} & 0.829 & 0.709 & 0.600
& 0.480 & 0.794 & 0.682 & 0.544 \\
& \cmark & 0.498 & \textbf{0.850} & \textbf{0.739} & \textbf{0.601}
& \textbf{0.487} & \textbf{0.829} & \textbf{0.717} & \textbf{0.553} \\
\cmidrule(lr){1-10}
\multirow{2}{*}{[20\%, 30\%)}
& \xmark & \textbf{0.678} & 0.913 & 0.852 & 0.743
& \textbf{0.560} & 0.833 & 0.723 & 0.600 \\
& \cmark & \textbf{0.678} & \textbf{0.935} & \textbf{0.872} & \textbf{0.746}
& 0.555 & \textbf{0.850} & \textbf{0.741} & \textbf{0.607} \\
\cmidrule(lr){1-10}
\multirow{2}{*}{[30\%, 40\%)}
& \xmark & \textbf{0.766} & 0.942 & 0.895 & 0.801
& \textbf{0.616} & 0.873 & 0.790 & 0.680 \\
& \cmark & \textbf{0.766} & \textbf{0.947} & \textbf{0.902} & \textbf{0.803}
& 0.614 & \textbf{0.878} & \textbf{0.806} & \textbf{0.686} \\

\midrule
\multicolumn{10}{c}{\textbf{FusionESP}} \\
\midrule
\multirow{2}{*}{$<$ 20\%}
& \xmark & \textbf{0.470} & 0.839 & 0.756 & 0.608
& 0.369 & 0.750 & 0.670 & 0.510 \\
& \cmark & 0.460 & \textbf{0.852} & \textbf{0.774} & 0.608
& \textbf{0.406} & \textbf{0.778} & \textbf{0.708} & \textbf{0.552} \\
\cmidrule(lr){1-10}
\multirow{2}{*}{[20\%, 30\%)}
& \xmark & 0.617 & 0.878 & 0.830 & 0.713
& 0.455 & 0.807 & 0.731 & 0.575 \\
& \cmark & \textbf{0.626} & \textbf{0.891} & \textbf{0.846} & \textbf{0.723}
& \textbf{0.481} & \textbf{0.829} & \textbf{0.758} & \textbf{0.610} \\
\cmidrule(lr){1-10}
\multirow{2}{*}{[30\%, 40\%)}
& \xmark & \textbf{0.728} & 0.923 & 0.871 & \textbf{0.793}
& 0.566 & 0.871 & 0.810 & 0.686 \\
& \cmark & 0.717 & \textbf{0.925} & \textbf{0.874} & 0.788
& \textbf{0.574} & \textbf{0.873} & \textbf{0.817} & \textbf{0.695} \\
\bottomrule
\end{tabular}
\end{table*}

In our framework, retrieval augmentation is introduced to inject neighborhood context at inference time and improve robustness under low-sequence-identity evaluation. In this subsection, we analyze how performance changes with the number of retrieved neighbors $K$, together with the associated time and memory overhead. We also examine whether the same retrieval-then-aggregation strategy can serve as a general plug-and-play module for existing ESI predictors.

\subsubsection{Effect of the Number of Retrieved Neighbors}
Fig.~\ref{fig_re_k} shows the effect of inference-time enzyme-side retrieval with different numbers of retrieved neighbors $K$. Compared with the base model without retrieval, retrieval augmentation improves AUROC, AUPRC, and F1 in most settings. The main gains are obtained with a small number of neighbors, such as $K=5$, while larger $K$ values bring diminishing returns. This trend suggests that nearby enzymes provide useful contextual evidence, whereas more distant neighbors may introduce less relevant signals and dilute the retrieval benefit. The gains are generally more evident in low-identity regimes (\eg, $<20\%$), indicating that neighborhood evidence is particularly useful when the query enzyme is weakly represented by the training distribution. Recall decreases in some settings despite improvements in F1, suggesting that retrieval augmentation can make predictions more selective by reducing spurious positive assignments. Additional molecule-side retrieval results are provided in Fig.~S2 in the Appendix.

We also analyze the computational and storage overhead of retrieval augmentation. The per-query inference overhead mainly consists of FAISS search, additional forward passes for retrieved support pairs, and score aggregation. In practice, the dominant cost comes from the $K$ support-pair forward passes. The measured per-query latency is reported in Fig.~S3 in the Appendix. The storage overhead comes from the enzyme memory bank constructed from CLEAN-encoded features \cite{yu2023enzyme}. The index size scales linearly with the number of indexed enzymes. Because the retrieval database is decoupled from model training, it can be expanded independently when external enzyme collections are available. Overall, using $K=5$ provides a practical trade-off between performance and inference cost.

\subsubsection{Generalizability as a Plug-and-Play Module}
We next evaluate whether inference-time retrieval augmentation can improve existing ESI predictors without changing their architectures or retraining. We select OmniESI and FusionESP as representative competitive baselines and apply the same neighbor-retrieval and aggregation procedure to them at inference time. For both baselines, we use $K=5$ retrieved neighbors at inference time. 

Table~\ref{tab_re_plug} reports the results across identity splits on both datasets. Retrieval augmentation consistently improves AUROC and AUPRC for both baselines across identity splits, indicating that the benefit of inference-time retrieval is not specific to RAMMESI. F1 scores remain stable or improve in most settings, suggesting that retrieved neighborhood evidence mainly refines ranking quality without substantially disrupting classification performance. The improvements are most evident in the $<20\%$ identity regime, where test enzymes have the weakest homology to training enzymes. As sequence identity increases, the marginal benefit becomes smaller, consistent with stronger base-model performance in easier regimes. Overall, these results support retrieval augmentation as a plug-and-play inference mechanism for improving performance under low-identity ESI evaluation.

\subsection{Case Study}

\begin{figure*}[!t]
\centering
\begin{minipage}[t]{0.62\textwidth}
  \vspace{0pt}
  \centering
  \includegraphics[width=\linewidth]{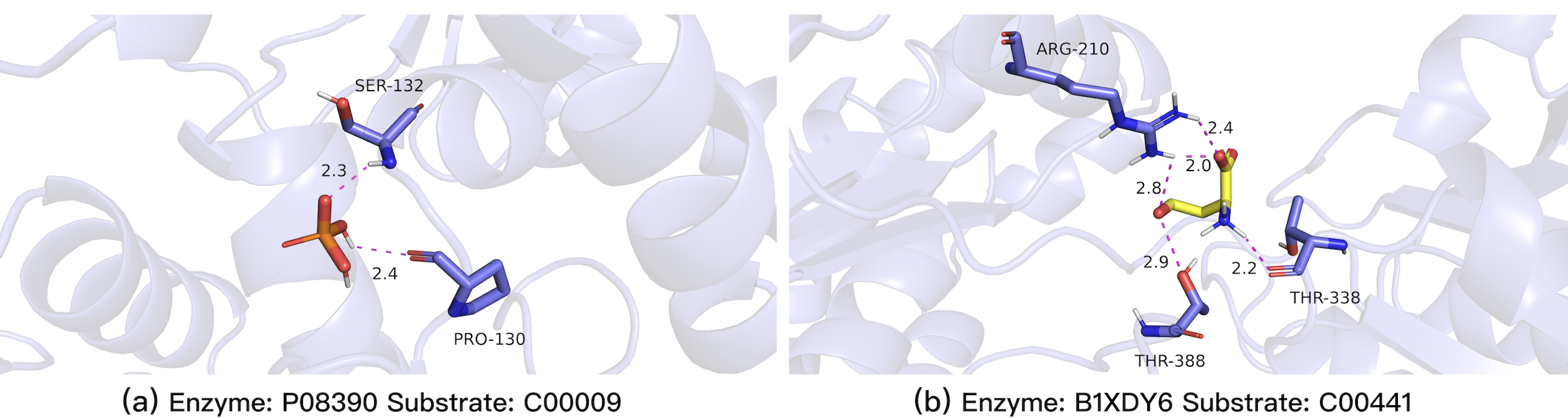}
  \caption{Docking visualization for two case-study pairs: (a) verified positive pair P08390-C00009; (b) high-scoring unlabeled candidate B1XDY6-C00441, supported by EC consistency and substrate similarity.}
  \label{fig_dock}
\end{minipage}
\hfill
\begin{minipage}[t]{0.36\textwidth}
  \vspace{0pt}
  \centering
  \captionof{table}{Case study performance.}
  \label{tab_case_study}
  \resizebox{\linewidth}{!}{%
    \begin{tabular}{lccc}
    \toprule
    \textbf{Method} & \textbf{Hits@10} & \textbf{Hits@100} & \textbf{EF@1.0\%} \\
    \midrule
    RandomForest & 2 & 5  & 1.78 \\
    LightGBM     & 0 & 2  & 1.42 \\
    DNN          & 0 & 2  & 2.49 \\
     ProSmith     & 0 & 4  & 3.20 \\
    VIPER        & 0 & 1  & 1.07 \\
    ESP          & 3 & 5  & 4.27 \\
    OmniESI      & 3 & 9  & 4.62 \\
    FusionESP    & 0 & 3  & 1.91 \\
    RAMMESI      & \textbf{6} & \textbf{16} & \textbf{7.12} \\
    \bottomrule
    \end{tabular}
  }
\end{minipage}
\end{figure*}

We next evaluate RAMMESI in a constrained-budget screening setting, where only a small fraction of candidate enzyme-substrate pairs can be experimentally assayed. We formulate this task as candidate prioritization: pairs are ranked by predicted interaction probability, and performance is measured by the recovery of verified interactions among top-ranked candidates. To construct the case dataset, we collect E.~coli enzyme-substrate interaction pairs from KEGG \cite{kanehisa2000kegg} and exclude enzymes appearing in either ESP-DB or Ryeactzyme-DB, ensuring an unseen-enzyme evaluation setting. The final dataset contains $90$ enzymes, $228$ substrates, and $20,520$ candidate pairs. Pairs without KEGG-curated evidence are treated as unlabeled candidates rather than confirmed non-reactive pairs. We report Hits@$K$ to count true positives among the top-$K$ candidates, and EF@1\% to measure enrichment over random selection within the top $1\%$ of the ranked list.

Table~\ref{tab_case_study} reports the screening results. RAMMESI achieves the best performance across all budgets, recovering $6$ true positives within the top $10$ candidates and $16$ within the top $100$, with the highest enrichment score (EF@1\% = $7.12$). At the stringent top-$10$ budget, only ESP and OmniESI retrieve true positives among baselines, whereas RAMMESI recovers twice as many as the strongest baseline. These results indicate that RAMMESI ranks reactive candidates more reliably for unseen enzymes under limited validation capacity.

To provide qualitative structural context for the screening results, we visualize two representative enzyme--substrate pairs and generate docking poses using AutoDock Vina \cite{trott2010autodock} (Fig.~\ref{fig_dock}). Here, docking is used only as supportive evidence for structural compatibility, rather than as proof of catalytic turnover. Fig.~\ref{fig_dock}(a) shows an experimentally verified positive pair, in which the docked substrate forms plausible pocket interactions with nearby residues, serving as a sanity check for the structural reasonableness of the ranking result. Fig.~\ref{fig_dock}(b) shows a high-scoring unlabeled candidate pair involving compound C00441 (L-aspartate-4-semialdehyde), which is not currently curated as a verified interaction. The corresponding enzyme is annotated as EC 1.2.1.41, whose native substrate ChEBI:58066 (L-glutamate-5-semialdehyde) is a close structural analog of C00441 (The 2D structures of C00441 and CHEBI:58066 are provided in Fig.~S4 in the Appendix). This functional consistency and chemical similarity suggest that the predicted pair is not arbitrary. In addition, the docking pose shows that C00441 can be accommodated in the pocket through multiple short polar contacts with residues such as ARG-210, THR-338, and THR-388, providing qualitative structural support for pocket compatibility. Overall, these examples suggest that RAMMESI can prioritize not only known reactive pairs but also structurally and functionally plausible unlabeled candidates beyond current annotations, which may be useful for downstream experimental screening.
 
\section{Conclusion and Discussion}
\label{conclusion}
This study presents RAMMESI, a retrieval-augmented multimodal learning framework for enzyme–substrate interaction prediction under low-homology shift. RAMMESI treats ESI prediction as a heterogeneous pairwise prediction problem, where reliable inference requires cross-modal compatibility modeling, learning from sparse positive supervision, and robustness beyond close-homolog transfer. The framework combines directional enzyme–substrate interaction modeling, pair-level joint representation learning, adaptive fusion, and imbalance-aware weighted-BCE optimization to capture substrate-specific reactivity from imbalanced supervision. To further improve robustness, RAMMESI introduces inference-time enzyme-side retrieval: neighboring enzymes are retrieved in representation space, recombined with the query substrate as support pairs, and scored by the trained pairwise predictor. The resulting predictions avoid directly transferring neighbor labels or retraining the base model.

Experimental results suggest that RAMMESI is effective particularly when ESI prediction cannot rely on close-homolog transfer. On the two open benchmarks ESP-DB and Reactzyme-DB, RAMMESI achieves consistently strong performance under sequence-identity-aware evaluation, especially in low-identity regimes. The retrieval analysis further indicates that inference-time neighborhood evidence improves ranking quality with limited overhead, and the plug-and-play experiments show that this benefit is not tied to a specific ESI backbone. In the constrained-budget screening case study, RAMMESI demonstrates its practical value for reaction prioritization by recovering more verified reactive pairs among top-ranked candidates and identifying structurally and functionally plausible unlabeled candidates.

From a knowledge and data engineering perspective, RAMMESI highlights a practical route for robust scientific pairwise prediction under sparse supervision and low-similarity distribution shift. In such settings, purely parametric models may be limited by incomplete positive observations and weak coverage of test entities, while direct nearest-neighbor transfer may ignore the pair-specific dependency that determines the target label. RAMMESI addresses this gap by combining learned multimodal compatibility with retrieval-based support evidence at inference time. Although this study focuses on ESI prediction, the same principle may be relevant to other heterogeneous interaction prediction tasks where candidate spaces are large, labels are incomplete, and predictions must generalize beyond closely related training examples.

Several directions remain open. First, the current retrieval memory is built mainly from single-side molecular representations. Future exploration may incorporate richer retrieval keys that encode interaction-level context, so that retrieved evidence is also selected by compatibility patterns between the two entities. Second, the role of retrieval and fusion under low similarity and high imbalance deserves further investigation beyond ESI prediction. Low similarity, high imbalance, and incomplete positive observations constitute a common data-engineering challenge in real-world heterogeneous pairwise prediction. In these settings, retrieval can provide non-parametric support from related examples, while fusion mechanisms determine how retrieved evidence should be integrated with parametric model predictions. Studying this interaction may extend the present framework from a specific ESI predictor to a more general strategy for robust pairwise prediction in scientific knowledge discovery.

\bibliographystyle{IEEEtran}
\bibliography{references}






\vfill

\end{document}